\documentclass[10pt,twocolumn,letterpaper]{article}

\usepackage{cvpr}
\usepackage{times}
\usepackage{epsfig}
\usepackage{graphicx}
\usepackage{amsmath}
\usepackage{amssymb}

\usepackage[breaklinks=true,bookmarks=false]{hyperref}

\cvprfinalcopy 

\def\cvprPaperID{****} 

\begin{document}
	
	\title{Language Independent Single Document Image Super-Resolution\\ using CNN for improved recognition}
	
	\author{Ram Krishna Pandey\\
		Indian Institute of Science\\
		Bangalore, India\\
		{\tt\small rkpandey@ee.iisc.ernet.in}
		\and
		A.G. Ramakrishnan\\
		Indian Institute of Science\\
		Bangalore, India\\
		{\tt\small ramkiag@ee.iisc.ernet.in}
	}
	\maketitle
	
\begin{abstract}
Recognition of document images have important applications in restoring old and classical texts. The problem involves quality improvement before passing it to a properly trained OCR to get accurate recognition of the text. The image enhancement and quality improvement constitute important steps as subsequent recognition depends upon the quality of the input image. There are scenarios when high resolution images are not available and our experiments show that the OCR accuracy reduces significantly with decrease in the spatial resolution of document images. Thus the only option is to improve the resolution of such document images.
The goal is to construct a high resolution image, given a single low resolution binary image, which constitutes the problem of single image super-resolution. Most of the previous work in super-resolution deal with  natural images which have more information-content than the document images. Here, we use Convolution Neural Network to learn the mapping between low and the corresponding high resolution images. We experiment with different number of layers, parameter settings and non-linear functions to build a fast end-to-end framework for document image super-resolution. Our proposed model shows a very good PSNR improvement of about 4 dB on 75 dpi Tamil images, resulting in a 3\% improvement of word level accuracy by the OCR. It takes less time than the recent sparse based natural image super-resolution technique, making it useful for real-time document recognition applications.   
\end{abstract}
\section*{Keywords}
Convolution Neural Network; Document Images; Parametric-ReLU; PSNR; Optical Character Recognizer.
\section{Introduction}

Research in AI tackles the problems which humans can solve easily, but are difficult for machines to solve. Rather than hard-coding these tasks as computer instructions, a learning based approach is more sensible where computers train/learn from and adapt to the real-world examples, in a hierarchical manner, from simpler to more complex situations. This approach closely resembles the way a person acquires knowledge from the world to behave in an "expected" or "sensible" manner. An "intelligent" being perceives real world information through its senses; but it is difficult to formally provide such information to the computers in its raw analog form. Hence data is digitized and further processed to more concise and efficient-to-compute numeric-vector format (called as feature-vector or more concisely as features) to be handled by training/learning algorithms. 

The choice of representation of features has significant effect on the performance  of the machine-learning algorithms. So a lot of effort needs to be put in designing and hand-coding the features. Lately, multi-layer neural-network based learning techniques, also called "deep learning algorithms", are gaining popularity as one need not manipulate raw data to a great extent, and training the network itself takes care of generating features at different layers in a hierarchy of complexity and suitable to the task it is being trained for.

In this paper, we use such techniques to enhance the quality of low-resolution document-images for aiding the subsequent recognition by OCR. We train a Convolution Neural Network (CNN) to learn the mapping between low- and high- resolution example images and generate high-resolution images from the test dataset of low-resolution images. A fully connected neural network architecture does not take into account the spatial structure of images. For instance, it treats input pixels which are far apart and close together in exactly the same way~\cite{nielsen2015neural}. Therefore the usage of convolution neural network is necessary, which learns local patterns and takes into account, the spatial structure of feature maps ~\cite{nielsen2015neural}. It is also much easier to train a CNN than fully-connected dense architectures due to lesser number of model parameters. In a broader sense, convolution neural network allows computational models that are composed of multiple processing layers to learn representations of data with multiple levels of abstraction ~\cite{lecun2015deep}.

Convolution neural networks have layers of "3D filters". Each filter operates on all the outputs obtained from the filters of the previous layer corresponding to a particular neighborhood. The area of the neighborhood involved with reference to the input image increases with every subsequent layer. For example at first layer one filter may detect a single vertical line, and another might detect horizontal lines. The next layer now takes the features from the previous layer and can detect (corner) more complex features and so on. This is a basic view of CNN. 

Handcrafting features is difficult in the case of document images and therefore convolution neural network is used, which is a data-driven approach to learn the right filter bank for a particular dataset. Features are learned as the training progreses from simple to more complex ones with increasing depth. Increasing depth doesn't always guarantee improved performance, but in our case the use of filters of size $[1\times 1]$ at three consecutive layers after the first hidden layer, each followed by ReLU or PReLU non-linearity, gives better performance than the same experiment performed over a three layer architecture. Convolution layers' filters of size $[1\times 1]$, described in detail in later sections, aims to further enhance the non-linearity in the model and decreases the dimensionality of the feature space.

There are various factors, which affect the accuracy of an OCR on document images the most important of them being the image resolution or spatial resolution. We often have low-resolution images at our disposal, where the high resolution counterpart is not available, or the document images which were scanned at very low resolution and the original image got destroyed or lost. When images are scanned at a low resolution and the quality is low, it affects the OCR recognition and gives bad results in terms of character or word level accuracy. This motivates us to work on the crucial stage of document image enhancement before passing it on to OCR.

There are several techniques by which high resolution images can be obtained. Performance gain is obtained even in bi-cubic interpolated images over plain low-resolution images. But with this method we can't recover high frequency components ~\cite{parker1983comparison}. Classical reconstruction based techniques require multiple low resolution images to reconstruct a higher-resolution image. Sparse representation dictionary learning based method, used for image super-resolution is very slow. This technique involves the use of an over-complete dictionary i.e. $ D^{{d*K}} $ where $d$ is the dimension of the signal and $ K \ge d $ ~\cite{yang2010image} ~\cite{yang2012coupled}.
Thus, there is a need for an algorithm or architecture, which is useful in such scenarios. We have designed a five layer convolution neural network, which learns mapping from low- to high- resolution image patches. We have created a large dataset contaning 5.1 million LR and HR patch pairs of sizes $ 16\times 16 $ and $ 10 \times 10 $ respectively,  for training the model. 

The rest of the paper is organized as follows. Related work is discussed in Section 2. Section 3 describes the problem formulation. Training model and dataset creation are explained in Sections 4 and 5, respectively.  Section 6 summarizes the experiments performed and results obtained. Finally, conclusions are drawn and concluding remarks are drawn in Section 7.

\section{Related work}
The technique for image super-resolution (SR) in the literature can be classified broadly into two categories: a classical multiple image SR and single image SR. The former approach requires multiple images with sub-pixel alignment to obtain a high resolution image~\cite{yang2010_1_image}. The latter approach is discussed in detail in ~\cite{nasrollahi2014super} and ~\cite{yang2014single}. The sparse-coding based method and its several improvements ~\cite{yang2010image}, ~\cite{yang2012coupled} are among the state-of-the-art SR methods. Chao Dong et. al.~\cite{dong2014image} proposed deep learning based natural image super-resolution, which showed that traditional SR method can be viewed as a deep convolution neural network which is light weight and achieves state-of-the-art restoration quality. It is also fast enough for practical applications.

\section{Problem formulation}
Given a low resolution image, the objective is to increase the resolution so that the OCR recogniton accuracy improves. The problem can be mathematically formulated as follows.\\ Assume that the given low resolution images is $ y $, and its bicubic interpolated image is $ Y $. The task is to learn mapping function $J$ to recover a high resolution image from $ Y $  i.e. $J(Y)$ which gives better OCR recognition accuracy than $Y$ and also improved PSNR.

The training set is defined as $I = \{(Y_{i},X_{i}) : 1 \leq i \leq N\}$, where $Y_{i}$ is the LR image and $X_{i} \approx J_{\lambda}(Y_{i}) $ is the corresponding HR image patch of the training set.
 
We use a parametric model to learn a non-linear mapping function $J_{\lambda}(Y)$, $\lambda$ being the parameters of the model, which minimize the reconstruction error between $J_{\lambda}(Y)$ and the corresponding ground truth high resolution image $X$. The learned model/function should best fit training data and generalize well to the test data.

The model is represented as,
 $\lambda = \{ W_{i}, b_{i} \} $ where
 $ W_{i} $ = $ \big\{ W_{i}^{j} : \hspace{.1cm} 1\leq j \leq n_{i}\big\} $, and $W_{i}^{j}$ is the $j^{th}$ filter at the $i^{th}$ layer as mentioned in Table~\ref{CNNarch}
and we learn these model parameters with a training algorithm in a supervised-manner as follows. \\

Let $ * $ denotes the convolution operation and $ 0 < \alpha < 1 $ is the data dependent parameter.\\
Initialize $ J_{0} = Y $ \\

$ \boldsymbol{for} \hspace{.2cm} i=1:4 \hspace{.2cm} \boldsymbol{do} $ \\

$ \hspace{1cm} Z^{j}_{i} =W^{j}_{i}*J_{i-1} + b_{i}$\\

$ \hspace{1cm}J^{j}_{i}= \max(Z_{i}^{j} , 0)\hspace{.5cm} \hspace{1cm} \bf (for\hspace{.1cm}ReLU) $
$$\boldsymbol{or}$$
$\hspace{1cm} J_{i}^{j} = \left \{ 
\begin{tabular}{c}
$\hspace{.0000001cm}\alpha Z_{i}^{j} \hspace{.3cm} if \hspace{.3cm} Z_{i}^{j} \le 0 $ \\

$\hspace{.1cm} Z_{i}^{j} \hspace*{1.1cm} else $ 
\end{tabular} 
\hspace{.5cm} \bf ( for\hspace{.1cm} PReLU)
\right\}$

$\boldsymbol{end} \hspace{.1cm} \boldsymbol{for} $

\vspace{.1cm}
$J_\lambda(Y) = W_{5} $ * $ J_{4}(Y) +  b_5 $\\



The loss function is given by,
\[L(\lambda)=\frac{1}{N}\sum\limits_{i=1}^{N}\parallel J_{\lambda}(Y_i)-X_i\parallel^2_{F}\]

To minimize this loss function, back-propagation with stochastic gradient descent algorithm (SGD) with momentum is used. The momentum provides faster convergence and reduced oscillation about the minima.  Stochastic gradient decent algorithm performs parameter update at each iteration and is faster than batch gradient descent(BGD). BGD performs redundant computations for large datasets, since it recomputes gradients for similar examples before each parameter update ~\cite{ruder2016overview}. SGD removes this redundancy by performing one update over each mini-batch.

\section{CNN model}
\subsection{Initializing the five layer network}
Deep neural networks have a highly non-linear architecture and the solution for the model parameters is non-convex. Initialization is one of the most important steps in designing deep neural models. If not performed with proper care, network may not perform well, and can even become dead or unresponsive in the middle of training, resulting in poorly trained model. Initialization means setting the model weights with initial values. Several initialization strategies have been proposed in the literature, some of them are : uniform initialization scaled by square root of the number of inputs i.e. W $ \sim U [\frac{-1}{\sqrt{n_{i}}} , \frac{1}{\sqrt{n_{i}}}  ] $ ~\cite{backproplecun} ; Glorot uniform initialization i.e. W $ \sim U [\frac{-1}{n_{i} + n_{o}} , \frac{1}{ n_{i} + n_{o} }  ] $, Glorot normal initialization i.e. W $ \sim \mathcal{N} [ 0 , \frac{1}{ n_{i} + n_{o} }  ] $ ~\cite{glorot2010understanding} and He-uniform initialization i.e, W $ \sim U [\frac{-1}{ n_{i} } , \frac{1}{ n_{o} }  ] $ ~\cite{he2015delving}, where $n_{i}$ and $n_{o}$ are the fan-in and fan-out of the  layer W respectively.

The paper ~\cite{glorot2010understanding} gives insights on the initialization of deep neural networks. In their work, they have explained that proper initialization helps signals to reach deep into the network. If the starting weights in a network are very small, then the signal shrinks as it passes through each layer until it is too small to be useful. However, if the starting weights in a network are large, then the signal grows as it passes through each layer until it is too large to be useful. All the initialization techniques mentioned above aims at keeping the weights in a controlled range during training.

For implementation reasons, it is difficult to find out how many neurons in the next layer consume the output of the current ones. In our work, we have used initialization presented in ~\cite{he2015delving}. It started from Glorot and Bengio~\cite{glorot2010understanding} and suggest using var(W)=${2}/{n_i}$  instead of ${2}/{(n_o+n_i)}$ ~\cite{glorot2010understanding}. This makes sense because a rectifying linear unit is zero for half of its input and hence we need to double the weight variance to keep the signal's variance constant.

\subsection{Non-linearity layers' activation function}
We use non-linear layers at several places in the CNN model. Among the non-linearities proposed, the most popular ones are sigmoid and rectified linear unit (ReLU) ~\cite{glorot2011deep}. Recently, extensions of ReLU non-linearities were proposed such as Leaky ReLU~\cite{leaky}, parametric ReLU (PRELU)~\cite{prelu}, SReLU~\cite{srelu} and randomized ReLU~\cite{rrelu}. In our experiments, we have used ReLU and PReLU units as activation functions. The advantage of using ReLU instead of sigmoid is because we avoid the vanishing gradient problem in the former case. ReLU is represented mathematically as $f(x)=max(0,x)$, i.e., it clips the -ve input to zero. ReLU can be implemented simply by thresholding a matrix of activations at zero and thus it is computationally much more efficient than sigmoid or tanh. Convergence of stochastic gradient descent with ReLU is  faster than for the sigmoid/tanh functions ~\cite{ruder2016overview}, due to its linear, non-saturating form. The disadvantage with ReLU is that it can be fragile during training and can "die" in the middle of training. For example, a large gradient flowing through a ReLU neuron may result in update of weights such that the neuron will never activate on any datapoint again. If this happens, then the gradient flowing through the unit will be zero forever from that point on ~\cite{li2015cs231n}. In other words, the ReLU units can irreversibly die during training since they can get knocked off the data manifold ~\cite{li2015cs231n}. This problem can be solved if the learning rate is set properly.

Other activation functions that can be used are Leaky ReLU, which reported better performance; but the results are however, not always consistent. It is the same as ReLU but the slope of the negative part is $0.01$. Parametric ReLu (PReLU) is also the same as ReLU, but the slopes of negative input are learned from the data rather than being pre-defined. In randomized ReLU (RReLU), the slopes for negative input are randomized in a given range while training, and then fixed during testing. It has been reported that RReLU can reduce over fitting due to its random nature ~\cite{xu2015empirical}.

\begin{table}[!h]
	\centering
\caption{5-Layer CNN architecture}
\label{table:3}
	\resizebox{0.47\textwidth}{!}
{
		\begin{tabular}{|c|c|c|c|c|c|} 
			\hline
			\textbf{Layer} & \textbf{Feature map} & \textbf{Filter} & \textbf{stride} & \textbf{pad} & \textbf{output} \\ [0.5ex] 
			\hline
			conv1 & $ 16\times16\times1 $ & $ 5\times5\times1\times64 $ & 1 & 0 & $ 12\times12\times64 $  \\ 
			\hline
			Act.fn. & $ 12\times12\times64 $ & $ ReLU/PReLU $ & 1 & 0 & $ 12\times12\times64 $ \\ 
			\hline
			conv2 & $ 12\times12\times64 $ & $ 1\times1\times64\times44 $ & 1 & 0 & $ 12\times12\times44 $ \\ 
			\hline
			Act. fn. & $12\times12\times44$ & $ ReLU/PReLU $ & 1 & 0 & $12\times12\times44$ \\ 
			\hline
			conv3 & $12\times12\times44 $ & $ 1\times1\times44\times24 $ & 1 & 0 & $ 12\times12\times24 $ \\ 
			\hline
		    Act.fn. & $12\times12\times44 $ & $ ReLU/PReLU $ & 1 &  0 & $ 12\times12\times24 $\\ 
			\hline
			conv4 & $ 12\times12\times24 $ & $ 1\times1\times24\times14 $ & 1 & 0 & $ 12\times12\times14 $ \\ 
			\hline
			Act.fn. & $ 12\times12\times14 $ & $ ReLU/PReLU $ & 1 &  0 & $ 12\times12\times14 $\\ 
			\hline
			conv5 & $ 12\times12\times14 $ & $ 3\times3\times14\times1 $ & 1 & 0 & $ 10\times10\times1 $ \\ [.1ex]
			\hline
		\end{tabular}}
		\label{CNNarch}
	\end{table}
	
\begin{figure}[!hr]
		\includegraphics[width=0.47\textwidth,height=0.32\textwidth]{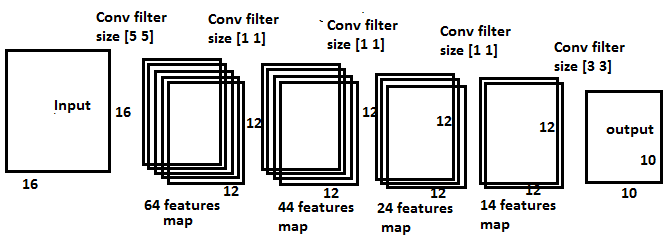}
		\caption{The 5-layer convolution neural network model with activation in between the layers.}
		\label{fig-cnn-model}
\end{figure}

Figure~\ref{fig-cnn-model} gives the architecture of the 5-layer CNN used for obtaining the HR patches from the LR input patches. The first layer has 64 filters each of size $[5\times 5]$ producing 64 feature maps of size $[12\times 12]$, for a $16\times 16$ image input patch. To further embed non-linearity into the CNN model and increase depth of the network to reduce the dimensionality of feature space, we use layers with filter size $[1\times 1]$. The use of filters of size $[1\times 1]$ in 3 consecutive layer after the first hidden layers followed by ReLU/PReLU activation to provide more non-linearity, helps to obtain good features. This model gives better performance than a plain three layer architecture.

From experiments, we found that using 2/3 times the number of filters as that of the previous layer gives better results.

\begin{figure}[!h]
	\includegraphics[width=0.49\textwidth,height=0.32\textwidth]{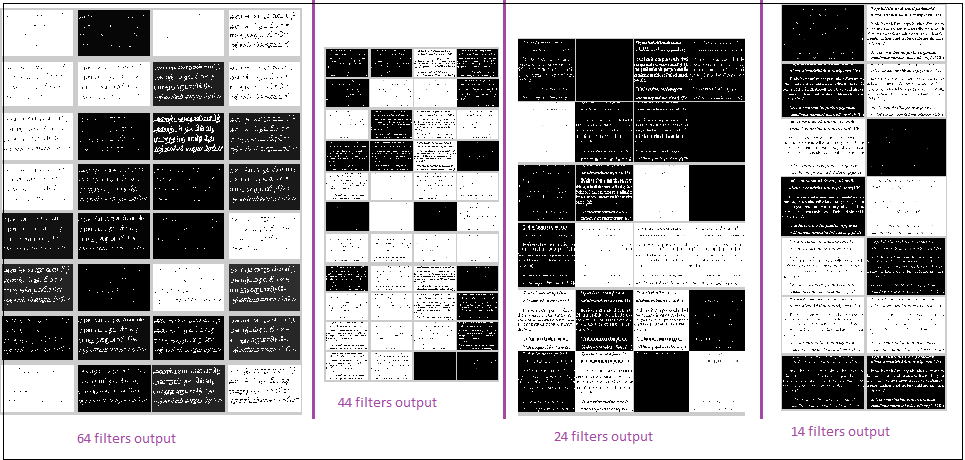}
	\caption{Output at each convolution layer showing the feature maps obtained at each layer for a sample input.}
	\label{fig-feat-maps}
\end{figure}

\begin{figure}[!h]
	\includegraphics[width=0.47\textwidth]{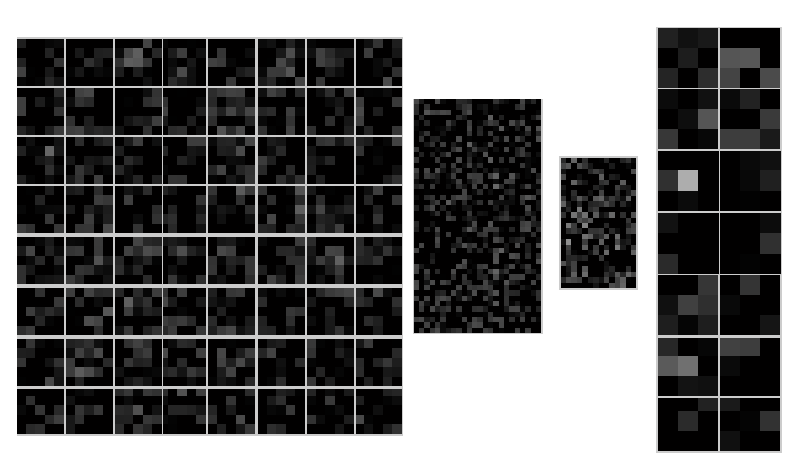}
	\caption{Learned filters at each layer which are used to obtain the feature maps~\ref{fig-feat-maps}}
	\label{fig-filters}
\end{figure}

Figure~\ref{fig-feat-maps} shows that the learned filters~\ref{fig-filters} are performing operations on the input image to obtain simple to more complex features at progressive layers. Since document images contain less number of features than natural images, we are reducing the number of features by using filters of size $[1\times 1]$  at three consecutive layers.

The input to CNN is a low resolution image of size $ [16\times16\times1]$. The first convolution layer computes the outputs of neurons that are connected to the local regions in the input, each computing a dot product between their weights and a small region they are connected to. In our case, we have used 64 filters of size $ [5\times5] $, which result in output volume of size $[12\times12\times64]$. Similarly, the second layer has 44 filters of size $[1\times 1]$ and generates output volume of size $[12\times12\times44]$. The third layer has 24 filters of size $1\times1$ which generates output volume of size $[12\times12\times24]$; and fourth layer has 14 filters of size $1\times1$ which generates output volume of size $[12\times12\times14]$. The last layer has 1 filter of size $3\times3$ to produce the output volume $[10\times10\times1]$. Each convolution layer is followed by a ReLU/PReLU non-linearity layer. ReLU/PReLU layer applies element-wise activation function, $max(0,x)$, and leaves the size of the volume unchanged. 

If the input to a convolution layer is of size $width1\times height1 \times depth1$ and at each layer we have four hyper-parameter namely number of filters (nf), filter spatial extent (se), stride (s) and amount of zero padding (zp); the output size $width2 \times height2 \times depth2$ is calculated according to the below mentioned formula~\cite{li2015cs231n}.

$$width2=(width1-se+2\times zp)/s+1$$
$$height2=(height1-se+2\times zp)/s+1$$
$$depth2=nf$$

Figure ~\ref{fig-feat-maps} shows the obtained feature maps at each convolution layer using the learned filters in Figure ~\ref{fig-filters} 
\section{The training and test datasets}

\begin{figure}[!h]
	\centering
	\includegraphics[width=0.25\textwidth]{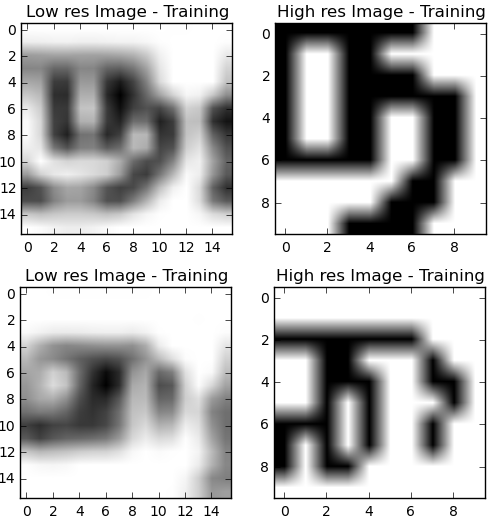}
	\caption{Some LR and HR pairs from the Tamil training image}
	\label{fig-dataset}
\end{figure}

We create a training dataset of 5.1 million HR-LR image patch-pairs, by randomly cropping from 135 document images. We use document images having text in three different languages, viz. Tamil, Kannada and English, scanned at different resolutions of 100, 200 and 300 DPI. This enhances robustness of the trained model towards input of different languages and resolutions. The Figure ~\ref{fig-dataset} shows some training samples of Tamil images.\\  The HR image patch is of size $10 \times 10$, and cropped directly from the document images. The corresponding LR patch is of size $16 \times 16$ and cropped from the image obtained by down-sampling and up-sampling the original document image with a factor of 2. The image patch-pair sizes smaller than the selected ones do not result in any significant performance improvement.

For the testing dataset, we use different set of 18 document images and generate HR-LR image patch-pairs in the same manner.




\section{Results of super-resolution experiments}
\begin{figure}[!h]
	\includegraphics[width=0.49\textwidth]{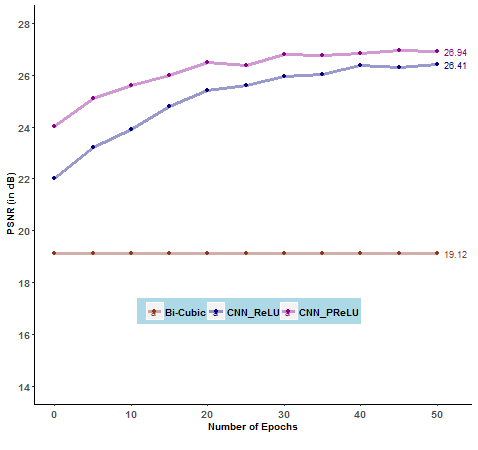}
	\caption{Training curve obtained using input images from three languages namely English, Tamil and Kannada at 150 dpi using ReLU or PReLU activations.}
	\label{tc}
\end{figure}

We implement the model using theano~\cite{theano} and Keras~\cite{keras} libraries in python. We used NVIDIA TITAN GTX GeForce (12GB memory) GPU for training.
The training image patches are of size $16\times16$, which are normalized by subtracting the mean over the entire training data and range-normalized to $[0,1]$, as pre-processing step. The training dataset is also randomly shuffled to prevent the model from being trained to some arbitrary data pattern. We have used MSE loss function and SGD solver with standard back-propagation. We trained for 50 epochs with the learning rate of 0.0001 and batch-size of 32. Figure ~\ref{tc} shows the training curve obtained using input images from three languages namely English, Tamil and Kannada at 150 dpi using ReLU and PReLU activations.

Suppose R is the reconstructed image and $ X_{h} $ is the ground truth image. We use PSNR (peak-signal-to-noise ratio) as the metric for image quality listed in Table ~\ref{PSNR}, which is calcutaed as follows:
$$E=  X_{h} - R$$
$$RMSE = \sqrt{\frac{1}{N^2}\parallel E \parallel_{F}^2}$$
$$PSNR = 20 log_{10}\frac{255}{RMSE}$$
\begin{figure}[!h]
	\includegraphics[width=0.49\textwidth,height=0.49\textwidth]{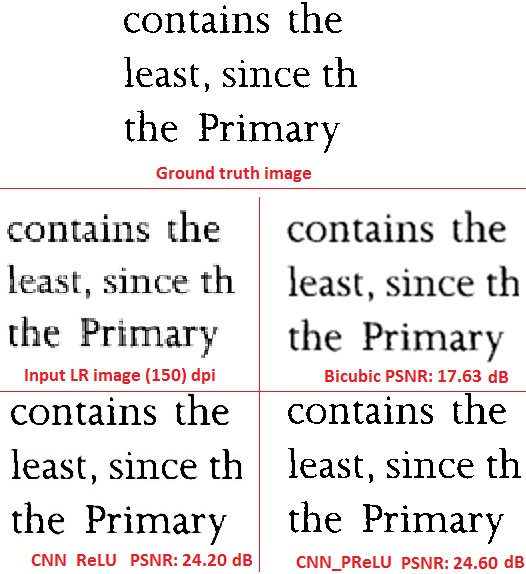}
	\caption{A sample English image with its improvement in PSNR over bicubic interpolation}
	\label{English}
\end{figure}
\begin{figure}[!h]
	\includegraphics[width=0.49\textwidth,height=0.49\textwidth]{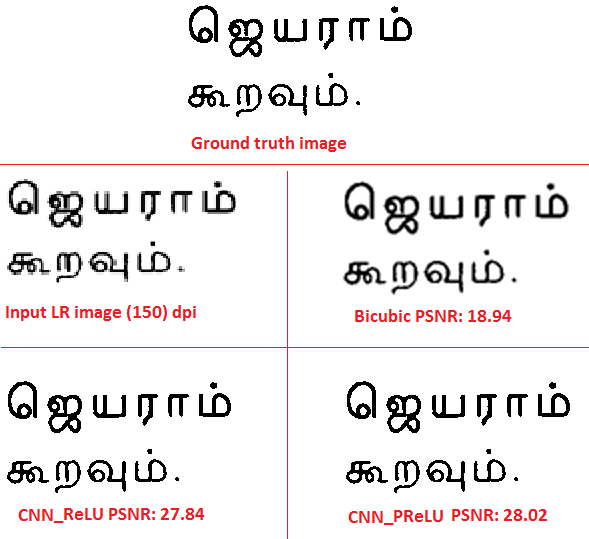}
	\caption{A sample Tamil image with its improvement in PSNR over bicubic interpolation}
	\label{Tamil_image}
\end{figure}

\begin{figure}[!h]
	\includegraphics[width=0.49\textwidth,height=0.49\textwidth]{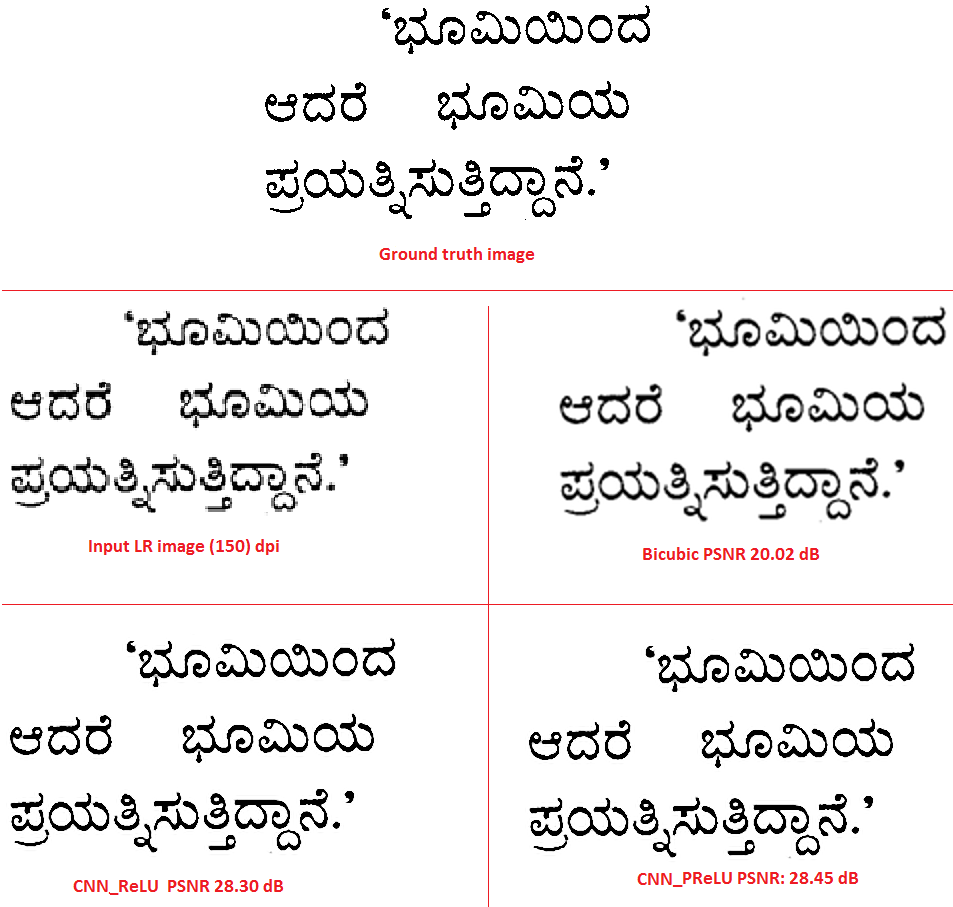}
	\caption{A sample Kannada image with its improvement in PSNR over bicubic interpolation}
	\label{Kannada_image}
\end{figure}

\begin{table}[!h]
	\begin{center}
		\caption{Comparison of average PSNR SR images with that of the bicubic interpolated images for each language for different input resolutions.}
		\resizebox{0.47\textwidth}{!}
		{
			\begin{tabular}{|l|c|c|c|c|}
				\hline
				Language & Resolution & Bicubic PSNR(dB)& \bf CNN\_ReLU PSNR(dB) & \bf CNN\_PReLU (dB) \\
				\hline\hline
				English & 150 & 18.62 & \bf 26.02 & \bf 26.3 \\
				\hline
				Tamil & 150 & 20.34 & \bf 28.29 & \bf 28.48 \\
				\hline
				Kannada & 150 & 18.41 & \bf 24.92 & \bf 26.12\\
				\hline
				English & 100 & 16.44 & \bf 21.56 & \bf 21.64 \\
				\hline
				Tamil & 100 & 16.94 & \bf 22.42 & \bf 22.72 \\
				\hline
				Kannada & 100 & 16.24 & \bf 20.55 & \bf 21.00\\
				\hline
				\bf Tamil   & \bf 75  & \bf 15.40 & \bf 19.20 & \bf 19.44 \\
				\hline
				English & 50 & 13.37 & \bf 14.81 & \bf 15.08 \\
				\hline
				Tamil & 50 & 13.30 & \bf 14.56 & \bf 14.97 \\
				\hline
				Kannada & 50 & 13.06 & \bf 14.46 & \bf 14.71\\
				\hline
				
			\end{tabular}
			\label{PSNR}
		}
	\end{center}
\end{table}

Table~\ref{OCR_accuracy}  lists the mean character level accuracy (CLA) and word level accuracy (WLA) obtained from the OCR performance as performance metrics for the quality of the HR image.

Figures [ ~\ref{English} ~\ref{Tamil_image} ~\ref{Kannada_image}] show comparison of results obtained on test images using CNN\_PReLU and ReLU over bicubic interpolated image.

\begin{table}
	\begin{center}
		\caption{Mean character and word level accuracies of CNN derived SR images (in \%) for 75 dpi Tamil input images.}
		\resizebox{0.40\textwidth}{!}
		{
			\begin{tabular}{|l|c|c|c|c|}
				\hline
				Metric & LR\_Input & Bicubic &  \bf CNN\_PReLU \\
				\hline\hline
				 CLA & 56.3 & 91.1 & \bf 94.0  \\
				 \hline
				 WLA & 12.5 & 59.6 & \bf 62.9 \\
				 \hline
			\end{tabular}
			\label{OCR_accuracy}
		}
	\end{center}
\end{table}

%
%





\section{Conclusion}

A five-layer CNN has been designed that can obtain high resolution document images from single low resolution images. The network's performance scales across the three languages that it has been tested for. Further, the same trained network works on different input resolutions of 50, 100 and 150 dots per inch. The increase in the PSNR value of the output image over a bicubic interpolated image varies from 1.5 dB to 7.8 dB for input images of 50 and 150 dpi, respectively. The OCR accuracy has improved by 3\% at the word level for Tamil images with 75 dpi input resolution. Thus, within the tested input resolutions and the languages, the proposed technique is language and resolution independent.\\
As future work, the performance of dedicated networks trained on a particular language and input resolution will be tested and compared with the results reported here.




{\small
\bibliographystyle{ieee}
\bibliography{egbib}
}

\end{document}